\documentclass[pmlr]{jmlr}

\RequirePackage{graphicx}
 \usepackage{booktabs}
 \usepackage{multirow}
 \usepackage{graphicx}

\usepackage{longtable}
\usepackage{colortbl} 
\definecolor{my-nose-color}{RGB}{219, 95, 87} 
\definecolor{head bottom}{RGB}{219, 140, 86} 
\definecolor{head top}{RGB}{219, 188, 88} 
\definecolor{left ear}{RGB}{205, 219, 87} 
\definecolor{right ear}{RGB}{205, 219, 87} 
\definecolor{left shoulder}{RGB}{111, 219, 87} 
\definecolor{right shoulder}{RGB}{86, 219, 109} 
\definecolor{left elbow}{RGB}{86, 219, 109} 
\definecolor{right elbow}{RGB}{87, 219, 203} 
\definecolor{left wrist}{RGB}{87, 188, 220} 
\definecolor{right wrist}{RGB}{87, 140, 219} 
\definecolor{left hip}{RGB}{87, 94, 218} 
\definecolor{right hip}{RGB}{126, 87, 219} 
\definecolor{left knee}{RGB}{173, 87, 218} 
\definecolor{right knee}{RGB}{219, 87, 218} 
\definecolor{left ankle}{RGB}{219, 87, 173} 
\definecolor{right ankle}{RGB}{219, 87, 126}

\makeatletter
\def\set@curr@file#1{\def\@curr@file{#1}} 
\makeatother
\usepackage[load-configurations=version-1]{siunitx} 

\theorembodyfont{\upshape}
\theoremheaderfont{\scshape}
\theorempostheader{:}
\theoremsep{\newline}

\jmlrvolume{298}
\jmlryear{2025}
\jmlrworkshop{Machine Learning for Healthcare}

\title[Newborn Screening Tool for General Movement Assessment]{Towards Scalable Newborn Screening: Automated General Movement Assessment in Uncontrolled Settings}

\author{\Name{Daphné Chopard}\thanks{Equal contribution}
       \Email{dchopard@inf.ethz.ch}\\ 
       \addr Department of Computer Science, ETH Zurich,
       Zurich, Switzerland \\ 
       \addr Department of Intensive Care and Neonatology and Children's Research Center, University of Zurich, University Children's Hospital Zürich, Zurich, Switzerland 
       \AND
       \Name{Sonia Laguna}{\footnotemark[1]}
       \Email{slaguna@inf.ethz.ch}\\ 
       \addr Department of Computer Science, ETH Zurich, Zurich, Switzerland
       \AND
       \Name{Kieran Chin-Cheong}{\footnotemark[1]}
       \Email{kchincheong@inf.ethz.ch}\\ 
        \addr Department of Computer Science, ETH Zurich, Zurich, Switzerland
       \AND
        \Name{Annika Dietz}
        \\
        \addr Department of Neonatology, University  Children’s Hospital Regensburg, Hospital St. Hedwig of the Order of St. John, University of Regensburg, Regensburg, Germany
       \AND
        \Name{Anna Badura}
        \\
        \addr Department of Neonatology, University  Children’s Hospital Regensburg, Hospital St. Hedwig of the Order of St. John, University of Regensburg, Regensburg, Germany
        \AND
        \Name{Sven Wellmann} 
        \\
        \addr Department of Neonatology, University Children’s Hospital Regensburg, Hospital St. Hedwig of the Order of St. John, University of Regensburg, Regensburg, Germany
        \AND
        \Name{Julia E Vogt} 
        \Email{julia.vogt@inf.ethz.ch}\\ 
        \addr Department of Computer Science,
       ETH Zurich,
       Zurich, Switzerland
       } 

\begin{document}

\maketitle
\vspace{-0.3cm}
\begin{abstract}
General movements (GMs) are spontaneous, coordinated body movements in infants that offer valuable insights into the developing nervous system. Assessed through the Prechtl GM Assessment (GMA), GMs are reliable predictors for neurodevelopmental disorders. However, GMA requires specifically trained clinicians, who are limited in number. To scale up newborn screening, there is a need for an algorithm that can automatically classify GMs from infant video recordings. This data poses challenges, including variability in recording length, device type, and setting, with each video coarsely annotated for overall movement quality. In this work, we introduce a tool for extracting features from these recordings and explore various machine learning techniques for automated GM classification.
\end{abstract}

\section{Introduction}

General movements (GMs) refer to spontaneous movements of the entire body observable in infants from early fetal life until approximately six months post-term~\citep{einspieler2005prechtl}. These movements include coordinates sequences of the arms, legs, neck, and trunk, with variations in intensity, force, and speed. GMs are inherently complex, variable, and fluid, and their quality can provide critical insights into the integrity of the developing nervous system. Assessing GM quality, known as the Prechtl General Movement Assessment (GMA), is a sensitive and reliable diagnostic tool for predicting the likelihood of cerebral palsy and other neurodevelopmental disorders \citep{einspieler2005prechtl}. The types of movements predictive of later neurodevelopmental disorders vary significantly with age, and age-based distinctions are essential when studying newborns. From early fetal life until approximately two months post-term, GMs exhibit consistent writhing movements. Between 6 to 9 weeks post-term, these writhing patterns diminish, and fidgety movements emerge, persisting until around the middle of the first year of life, at which point intentional movements become predominant. Alterations in these movement patterns can indicate later disorders. While GMA is typically conducted in hospitals by specialized clinicians, there are too few trained experts to screen all newborns. This motivates the need to develop a tool for automatic and reliable classification of movements. 

Initial attempts at addressing this challenge focused on computer-based approaches using feature extraction techniques \citep{baccinelli2020movidea, raghuram2022automated}. However, recent advances in deep learning and computer vision have spurred the development of more sophisticated methods \citep{silva2021future, irshad2020ai}. For instance, \citet{reich2021novel} proposed to use a pose estimator, OpenPose, to extract skeletons and used these as input to a shallow multilayer neural network to discriminate between movements. Other deep learning solutions, e.g. by \citet{schmidt2019general}, work in controlled conditions. Despite their promise, these approaches often rely on controlled environments with fixed sensors and constant recording length, limiting their scalability and applicability in real-world clinical settings.

In this work, we present a preliminary study introducing a method to label key anatomical points, automatically process videos, and classify GMs based on the tracked anatomical points across videos, using a dataset from the Barmherzige Brüder Regensburg Hospital.
The challenges posed by the variability in recording lengths and devices, the diversity of video scenarios (including both hospital and home environments), and the nature of the movement quality annotations---one per recording---are significant for reliable GM classification. Addressing these challenges is crucial to the generalization of an automatic GMA that can be effectively deployed in varied settings. The goal is to enable comprehensive newborn screening and follow-up studies by experts, ultimately improving the early detection and treatment of neurodevelopmental disorders.

\subsection*{Generalizable Insights about Machine Learning in the Context of Healthcare}

This work provides insight into the feasibility of building screening tools for GMA from videos captured in an uncontrolled environment. Unlike prior studies conducted in highly controlled environments, our dataset reflects clinical reality: varied recording devices, environments, and video quality. We show that even with coarse, single-label annotations per video---a common limitation in clinical data---robust classification of infant movement quality is achievable. Our findings emphasize the value of carefully designed preprocessing pipelines and simple, computationally efficient models, which perform competitively despite limited data. Additionally, the performance gap between early and late infancy groups suggests that age-specific modelling strategies may be necessary, highlighting the importance of aligning machine learning approaches with developmental context.



\section{Dataset}

\label{data}
This study includes 76 infant video recordings collected from the Barmherzige Br\"uder Regensburg Hospital. 
The dataset is highly heterogeneous, including videos recorded both at home and in hospital settings using various recording devices, resulting in differing resolutions and frame rates. There is also significant variability in camera angles and orientations, distance from subject to camera, infants' clothing, and background types, among other factors.

The data is divided into two age-based groups according to the GM phase. The \textit{early General Movement} group includes 39 preterm infants recorded after birth at postmenstrual ages between 32 and 36 weeks (mean: 33 weeks), while the \textit{late General Movement} group comprises 37 infants with postmenstrual ages ranging from 49 to 59 weeks (mean: 53 weeks). Each sample is associated with a binary label reflecting the infant's movement quality.  Following \cite{einspieler2005prechtl}, in the \textit{early GM} group, which encompasses preterm and writhing GMs, labels distinguish between normal and poor movement repertoire. In contrast, the \textit{late GM} group labels indicate the presence or absence of fidgety movements, with the presence of fidgety movements considered normal. 

To minimize bias and improve reliability, two trained physicians independently annotated the videos. In cases of disagreement, the recordings were reviewed jointly until a consensus label was reached.  In the dataset, 24 infants in the early GM group (65\%) exhibit poor repertoire, while 10 infants in the late GM group (27\%) lack fidgety movements. These proportions are consistent with previous clinical studies in high-risk populations \cite{saether2016change,de2017predictive,alonzo2022high}, although notably higher than the prevalence of about 3\% reported in cohorts of general population \cite{wu2020typical}. 

\begin{figure*}[h!]
    \includegraphics[width=15cm]{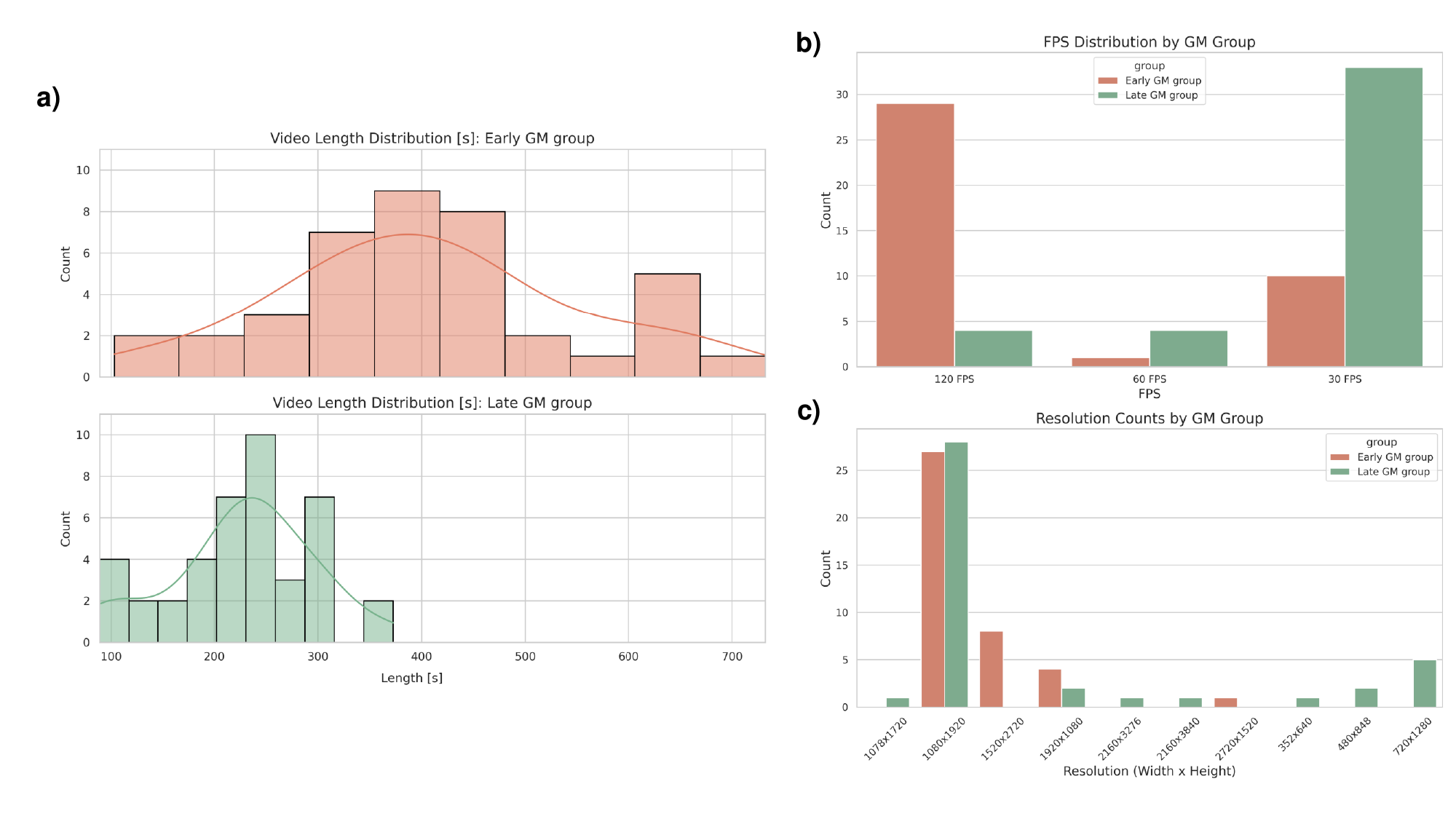}
    \vspace{-0.4cm}
    \caption{Histograms showing the distributions of (a) video length, (b) frame rate per second (FPS), and (c) video resolution for each GM group. The videos are highly heterogeneous, with substantial variation in recording length, frame rate, and resolution. FPS values are rounded to the closest one.}
    \label{fig:video_histograms}
\end{figure*}

\autoref{tab:dataset_characteristics} shows several demographic characteristics, as well as details about the videos themselves for both the early and late GM groups. Unfortunately, some details are not available to us, such as weight at video recording time for early GM group infants. \autoref{fig:video_histograms} provides histograms for the distributions of both video resolution and FPS for the early and late GM groups. Note that not all video segments are equally informative: while GMA usually requires 2 to 5 minutes of video, 15 to 30 seconds of relevant movement may be sufficient \cite{kapil2024unveiling}.

\begin{table*}[h!]
\caption{Demographic and video recording characteristics of infants in both GM groups.}
\centering
\resizebox{\textwidth}{!}{%
\begin{tabular}{ccc}
\toprule
\textbf{Characteristic} & \textbf{Early GM Group} & \textbf{Late GM Group} \\ 
\midrule
Age at video recording [days] (mean $\pm$~SD) & $30.9 \pm 19.9$ & 170.4 $\pm$ 24.8 \\
Post-menstrual age at video recording [days] (mean $\pm$~SD) & 231.5 $\pm$ 6.5 & N/A \\
Corrected Age at video recording [days] (mean $\pm$~SD) & N/A & 92.5 $\pm$ 17.4 \\
Weight at video recording [g] (mean $\pm$~SD) & N/A & 4148.2 $\pm$ 830.7 \\
Video Length [s] (mean~$\pm$ SD) & 409 $\pm$ 144 & 224 $\pm$ 93 \\
FPS [min, max] & [30, 120] & [30, 120] \\
Extremely Preterm & 16 & 15 \\
Very Preterm & 20 & 19 \\
Moderate to Late Preterm & 3 & 3 \\
Not Preterm & 0 & 0 \\

\bottomrule
\end{tabular}
}
\label{tab:dataset_characteristics}
\end{table*}

\section{Methods}

\begin{figure}[h]
    \centering
    \includegraphics[width=\linewidth]   
    {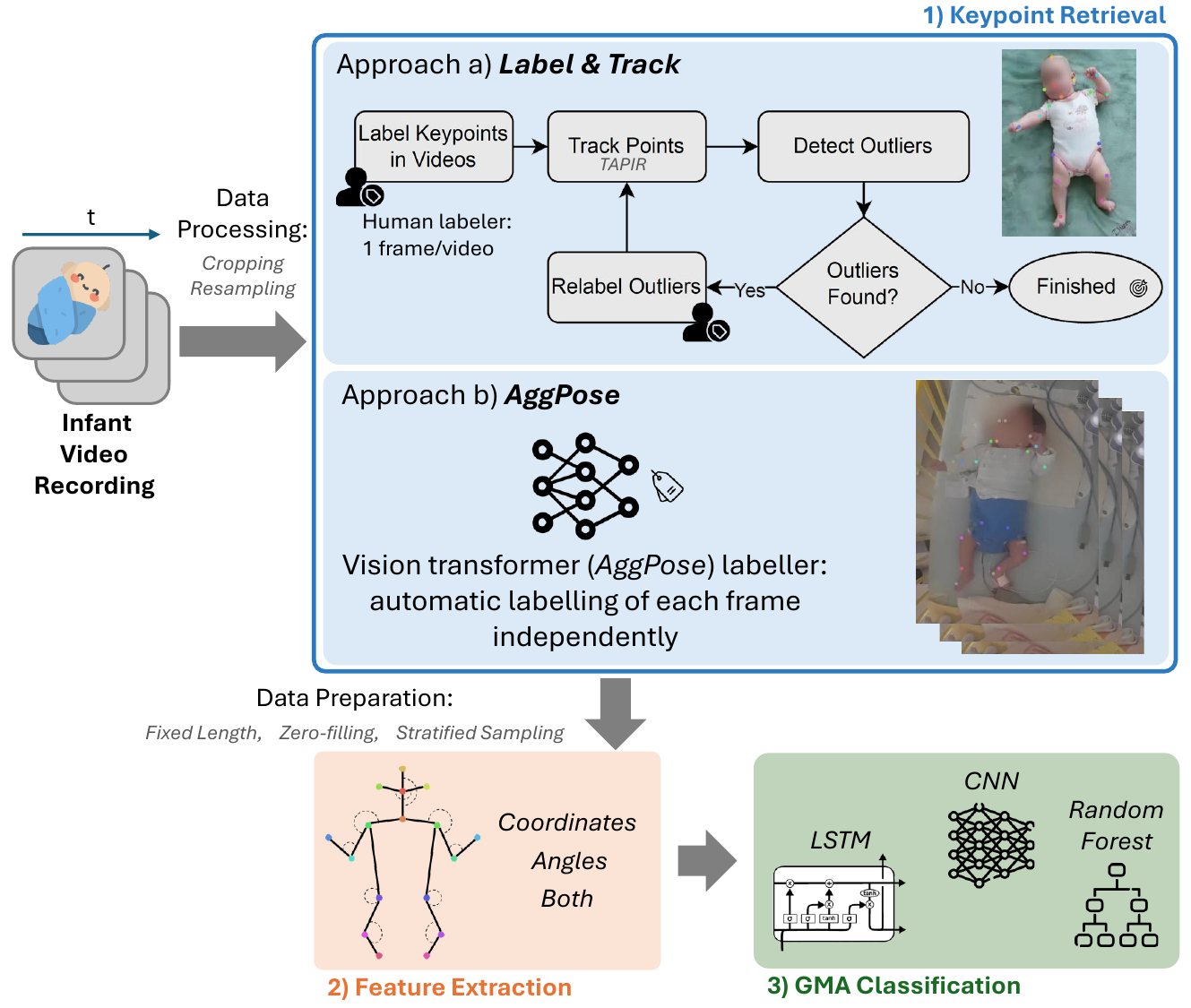}
    \vspace{-0.7cm}
    \caption{
    Overview of the GMA classification pipeline. Video recordings undergo preprocessing steps (Section~\ref{data_preprocessing}). Keypoints are then extracted using one of two approaches described in Section~\ref{sec:keypoint_ret}: (a) \textbf{Label \& Track}, or (b) \textbf{AggPose}. Features are extracted (Section~\ref{data_extract}) from the keypoints and used to train GMA classification models (Section~\ref{classifier_list}) including CNN, LSTM, and Random Forest for prediction.
    }
    \label{fig:preprocessing_pipeline}
\end{figure}

This work involves extracting features from specific anatomical landmarks of interest---so-called \textit{keypoints (KP)}---in video frames (Section~\ref{data_extract}) to classify movement quality. We explore two means of acquiring the desired keypoints: 1) \textbf{\textit{Label \& Track}}, manually labelling specific KPs in a video frame (Section~\ref{tools}), and tracking their positions across the video (Section~\ref{tools}), and 2) \textbf{\textit{AggPose}}, automatically labeling the KPs in each frame, using the existing ML labeller AggPose \citep{cao2022aggpose}. The overall pipeline is illustrated in Figure~\ref{fig:preprocessing_pipeline}. The \hyperlink{https://github.com/mdslabeth/GMA}{code}\footnote{\scriptsize \url{https://github.com/mdslabeth/GMA}} is publicly available.

\subsection{Keypoint Retrieval Tools}
\label{sec:keypoint_ret}
In this section, we provide details about the two approaches for extracting keypoints from videos. In the first approach, \textit{Label \& Track}, keypoints are manually labelled in a single frame and then tracked throughout the rest of the video. In the second approach, \textit{Automatic Label}, keypoints are automatically labelled in each video frame independently.

\label{tools}
\subsubsection{Label \& Track}
\label{tools1}
\paragraph{Keypoint Labelling} We introduce a keypoint labelling tool based on the work from \citet{doersch2023tapir} to mark sets of key coordinates in the videos, which, along with point tracking, forms the backbone of the preprocessing pipeline. The tool allows labelling of extremities such as wrists and ankles---referred to as \textit{extreme keypoints}, as well as additional joints such as elbows, hips, and knees---collectively referred to as \textit{all keypoints}, totaling 17. In this study, three independent non-expert labellers manually labelled all keypoints for subsequent tracking. Full lists of relevant keypoints can be found in~\autoref{fig:list_keys}.

\paragraph{Point Tracking} 
For tracking labelled keypoints across video frames, we utilize the TAPIR point tracking algorithm proposed by \citet{doersch2023tapir}.
TAPIR, is designed to track arbitrary physical points on surfaces over video sequences by combining per-frame initialization with temporal refinement.  It begins by computing a cost volume (namely a 4D map with dimensions corresponding to time, height, width, and channels) that captures the similarity between the feature representation of a query point and all possible spatial features in each frame. This cost volume is used to generate an initial estimate of the point's position, along with associated occlusion and uncertainty metrics. This is followed by an iterative refinement stage, where the local spatio-temporal features are processed using a depthwise-convolutional network to improve tracking accuracy over time.
TAPIR is trained on a modified version of the Kubric MOVi-E dataset~\cite{greff2022kubric}, which contains simulated videos featuring physically realistic interactions between deformable and rigid objects. Training supervision includes Huber loss for point location regression and binary cross-entropy loss for occlusion and uncertainty prediction. TAPIR is evaluated on TAP-Vid~\cite{doersch2022tap}, a collection of four diverse datasets (including TAP-Vid-Kinetics and TAP-Vid-DAVIS) each posing distinct tracking challenges. It achieves state-of-the-art performance and demonstrates strong robustness in occlusion-heavy and dynamic scenes. 

\begin{figure}[h]
    \centering
    \includegraphics[width=0.7\linewidth]{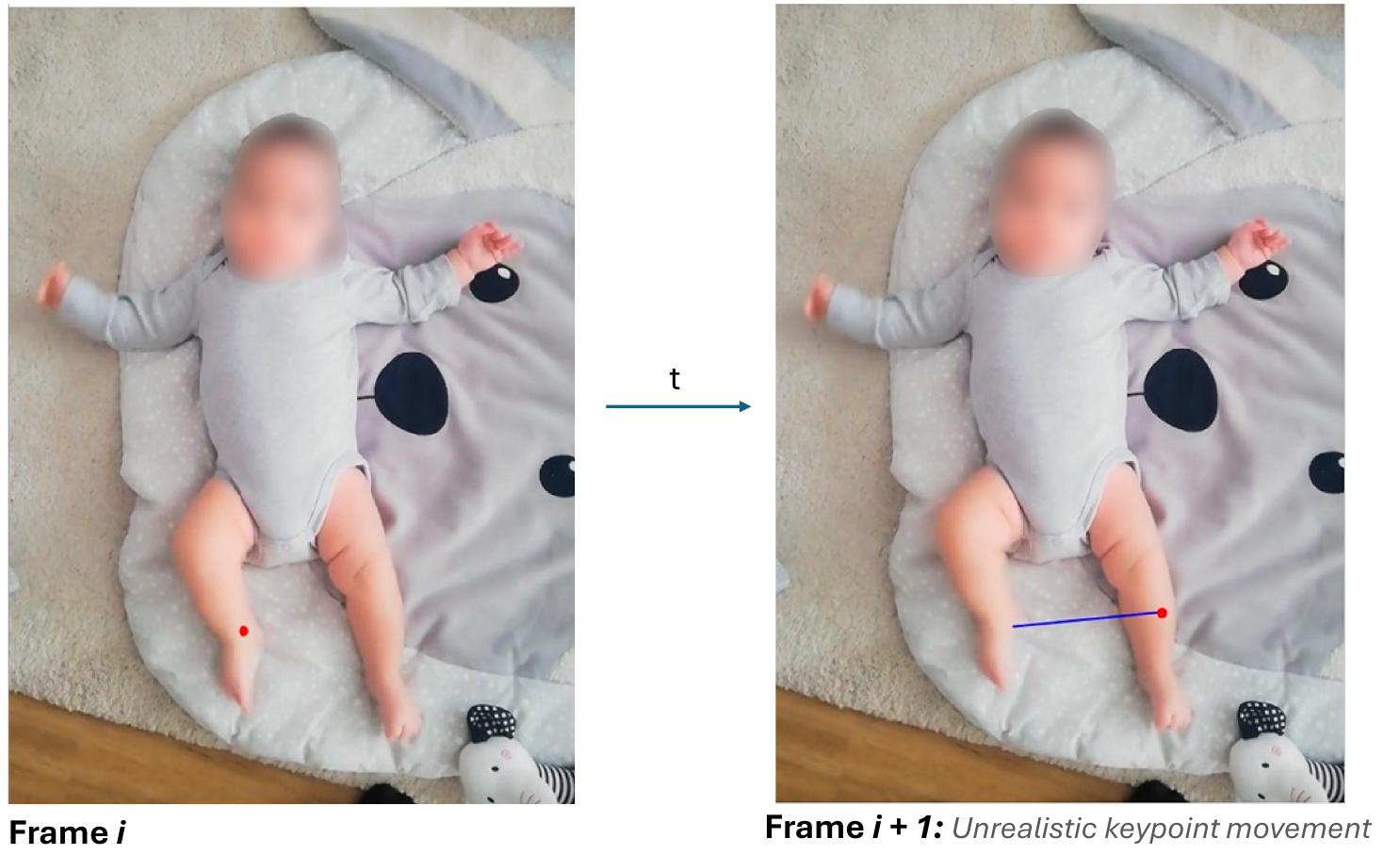}
    \caption{Consecutive frames with unrealistic keypoint tracking. The outlier detection tool displays potential unrealistic point movements and allows manual correction.} 
    \label{fig:tools_figure}
\end{figure}

\subsubsection{AggPose: Automatic label}

As a benchmark, we also evaluate a fully automated keypoint extraction method: AggPose \citep{cao2022aggpose}, a vision transformer model specifically designed for infant pose estimation and trained on a large-scale infant pose dataset. AggPose has been shown to outperform established models such as OpenPose \cite{cao2019openpose} and HRNet \cite{wang2020deep}, making it a strong and representative baseline for this class of methods. Unlike our \textit{Label \& Track} pipeline, AggPose eliminates the need for both manual labelling and tracking by independently predicting keypoints in each frame.

Unlike traditional convolutional models, AggPose discards convolutional layers in favour of a fully transformer-based architecture, leveraging multi-scale feature aggregation through a deep-layer fusion mechanism. This method enhances spatial information sharing across different resolution levels, improving the robustness of KP detection in challenging infant pose scenarios. We provide a comprehensive overview of this method in Appendix~\ref{app:aggpose}. 

In this study, we use AggPose to automatically label KPs in each frame of the video dataset, extracting the full set of coordinates across time. This method encompasses a total of 21 KPs, slightly larger than the prior approach. See \autoref{fig:preprocessing_pipeline} Approach b) for a visualization of the KPs, compared to manual labelling, some keypoints are omitted (head top, ears, and nose), while others are added (eyes, hand and foot extremities, torso)--displayed in grey--when labelling automatically.

\subsection{Data Preprocessing}
\label{data_preprocessing}

Due to the heterogeneity of the data, a robust preprocessing and error correction pipeline is necessary to prepare the videos for machine learning model development. This subsection outlines the key steps in that process, with a graphical overview in \autoref{fig:preprocessing_pipeline} Approach a. 

\paragraph{Resampling and Extreme Keypoints Labelling} Preprocessing begins with homogenizing all the video data by resampling to 30 frames per second. Extreme keypoints are then labelled. If all required keypoints are not visible in the first frame, labelling occurs in a subsequent frame where visibility is clearer.

\paragraph{Video Cropping} Next, the extracted extreme keypoints are tracked and used to apply a rectangular crop per video. The crop is determined by the most extreme values of the tracked points in each video, with a 15\% margin to ensure the infant consistently occupies a similar portion of the video frame in all videos. This normalization step is crucial for maintaining consistency in the input data for machine learning models.

\paragraph{Outlier Detection} To address inaccuracies in the tracking algorithm in the \textit{Label \& Track} (See \ref{tools1}) approach, such as when points jump inappropriately due to occlusions or overlapping limbs, an outlier detection mechanism is applied. 

The tool automatically flags any sudden and unrealistic movement of keypoints and prompts manual relabelling by the annotators, with an example shown in~\autoref{fig:tools_figure}. Specifically, we define such movements as coordinate changes that exceed 15 times the overall standard deviation for each keypoint. This threshold is set deliberately high to avoid flagging normal, natural movements, as well as fidgety movements that can inherently exhibit high magnitudes without indicating an error. After relabelling, the affected keypoints are retracked. This process is iterated a fixed number of rounds, two in these results, as we observed only marginal improvements in performance after three rounds in the final detection task, or until no further outliers are detected, ensuring the accuracy of the tracked coordinates. In general, this outlier-detection process can be structured to run for a set number of iterations, until a defined performance tolerance is reached, or until no further outliers are identified, ensuring the accuracy of the tracked coordinates. 

\paragraph{Final Data Preparation} Prior to classification, we homogenize the time series data. After resampling to 30 fps, each video is split into fixed-length sequences equal to the shortest recording within its respective age group (early: 616 frames; late: 674 frames; approximately 20.5 s and 22.5 s, respectively). Zero-filling is applied to any occluded or missing keypoint coordinates. To prevent data leakage, we use stratified sampling to ensure that no infant appears in both the training and testing datasets. If a video contains multiple full-length clips, each is treated as a separate instance; incomplete final fragments are discarded.

\subsection{Feature Extraction}
\label{data_extract}

We consider three feature sets from the tracked keypoints to perform the classification task: $x$- and $y$-coordinates of all keypoints and the angles between selected keypoints, based on \citet{prakash2023video}.

\begin{figure}[h!]
    \centering
    \begin{minipage}{0.4\linewidth} 
        \caption{List of \textit{all} and \textit{extreme} keypoints considered in this study for manual labelling, coloured by label on the right.}
        \label{fig:list_keys}
        \centering
        \begin{tabular}{p{2.103cm}|c} 
        \hline
        All Keypoints & Extreme \\
        \hline
        \rowcolor{my-nose-color} nose & \\
        \rowcolor{head bottom} head bottom & \\
        \rowcolor{head top} head top & \textbf{x} \\
        \rowcolor{left ear} left ear & \\
        \rowcolor{right ear} right ear & \\
        \rowcolor{left shoulder} left shoulder & \\
        \rowcolor{right shoulder} right shoulder & \\
        \rowcolor{left elbow} left elbow & \textbf{x} \\
        \rowcolor{right elbow} right elbow & \textbf{x} \\
        \rowcolor{left wrist} left wrist & \textbf{x} \\
        \rowcolor{right wrist} right wrist & \textbf{x} \\
        \rowcolor{left hip} left hip & \\
        \rowcolor{right hip} right hip & \\
        \rowcolor{left knee} left knee & \textbf{x} \\
        \rowcolor{right knee} right knee & \textbf{x} \\
        \rowcolor{left ankle} left ankle & \textbf{x} \\
        \rowcolor{right ankle} right ankle & \textbf{x} \\
        \hline
        \end{tabular}
    \end{minipage}%
    \hspace{0.15cm}
    \begin{minipage}{0.28\linewidth} 
        \centering
        \includegraphics[width=\linewidth]{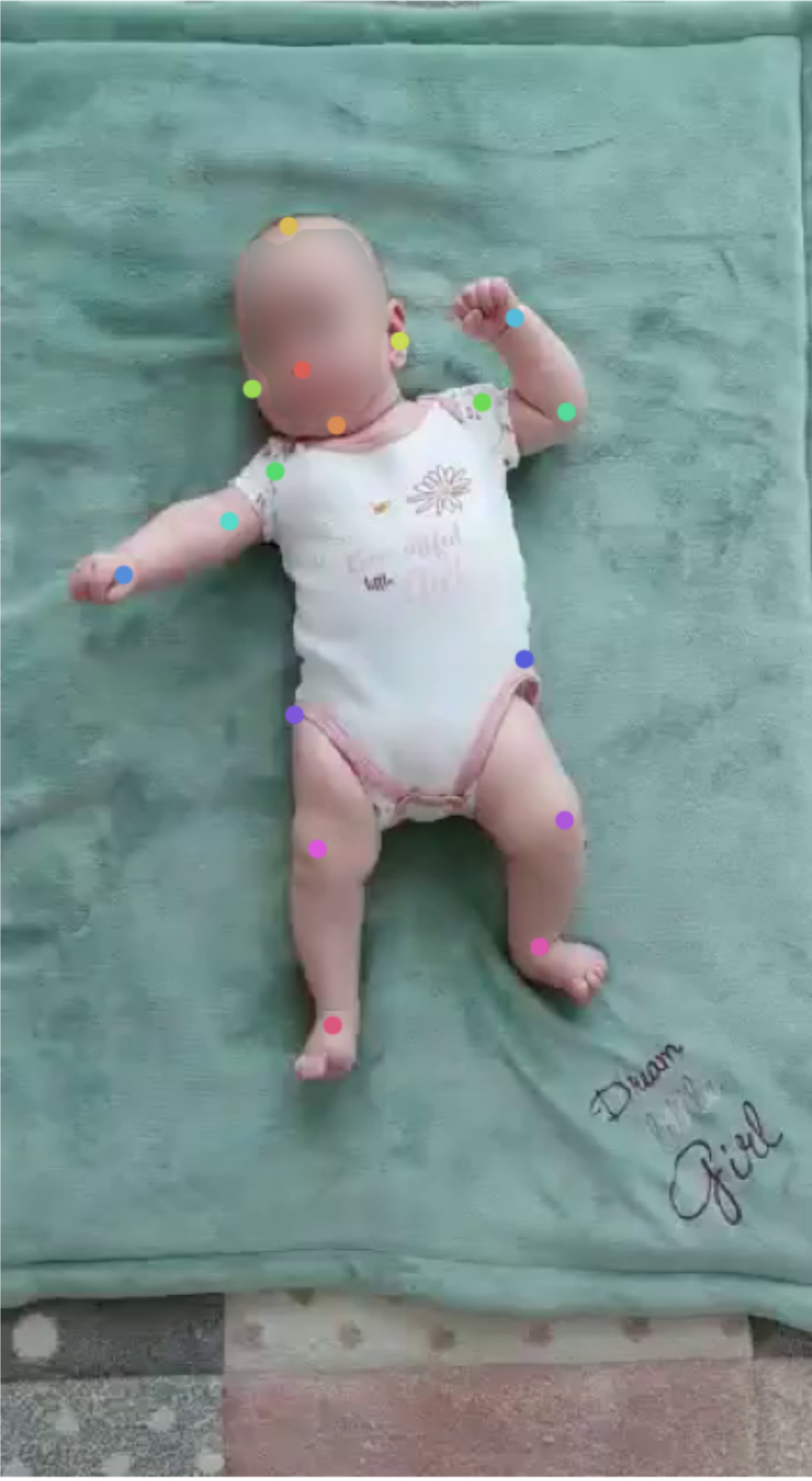} 
        \caption{Example of frame with labelled keypoints.}
        \label{fig:tools_figure_baby}
    \end{minipage}%
    \hspace{0.2cm}
    \begin{minipage}{0.28\linewidth} 
        \centering
        \includegraphics[width=\linewidth]{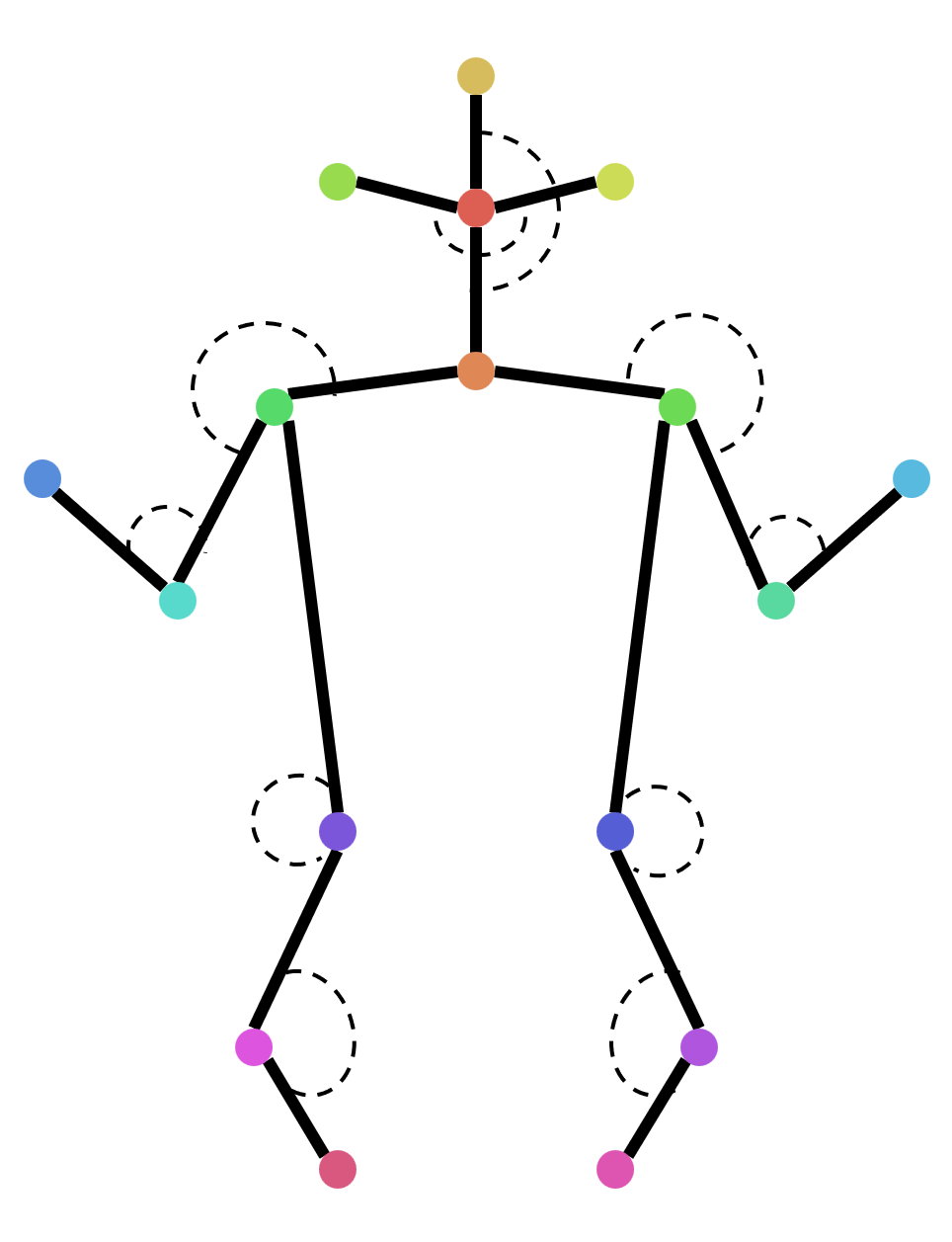} 
        \vspace{0.6cm}
        \caption{Illustration of the angle features. The dash lines show the angles that we compute in the set of angle features.}
        \label{fig:angle_features}
    \end{minipage}

    \label{fig:keypointlabelling}
\end{figure}

\paragraph{Keypoint Coordinates} The $x$- and $y$-coordinates of all labelled keypoints are treated independently as separate input channels, and used as a simple representation of the keypoint positions across frames, as listed in \autoref{fig:list_keys} and \autoref{fig:tools_figure_baby}. 

\paragraph{Keypoint Angles} The angles are calculated between keypoint triplets to capture movement dynamics. 
Given three keypoints $\mathbf{p}_1$, $\mathbf{p}_2$, and $\mathbf{p}_3$ with coordinates $(x_1, y_1)$, $(x_2, y_2)$ and $(x_3, y_3)$ respectively, we compute the angle $\theta$ formed at middle point $\mathbf{p}_2$ between vectors $\mathbf{v}_1 = \mathbf{p}_1 - \mathbf{p}_2$ and $\mathbf{v}_2 = \mathbf{p}_3 - \mathbf{p}_2$, defined as 

\begin{align}
\mathbf{v}_1 &= \begin{bmatrix} x_1 - x_2 \\ y_1 - y_2 \end{bmatrix}, \quad 
\mathbf{v}_2 = \begin{bmatrix} x_3 - x_2 \\ y_3 - y_2 \end{bmatrix} \\
\theta &= \arccos\left( 
\frac{\mathbf{v}_1 \cdot \mathbf{v}_2}
     {\|\mathbf{v}_1\| \|\mathbf{v}_2\|} 
\right).
\label{eq:angle_calc}
\end{align}

To ensure numerical stability, the cosine value is clipped to the range $[-1, 1]$ before applying the arccos function. 
Here, $\cdot$ denotes the dot product and $\|\cdot\|$ the Euclidean norm.

\autoref{tab:angle_pairs_explanation} lists the selected angle sets and the anatomical keypoints involved, with intuitive descriptions. These are illustrated in \autoref{fig:angle_features}. Although the automatic labelling method predicts more keypoints, the same set of angles is used.

\begin{table*}[h!]
\centering
\begin{tabular}{ccc|l}
\toprule
\textbf{$\mathbf{p}_1$} & \textbf{$\mathbf{p}_2$} & \textbf{$\mathbf{p}_3$} & \textbf{Explanation} \\ 
\midrule
head top   & nose         & head bottom     & Head top to neck angle \\ 
right ear  & nose         & left ear        & Right to left ear angle \\ 
left elbow & left shoulder & head bottom     & Head to left shoulder angle \\ 
right elbow & right shoulder & head bottom    & Head to right shoulder angle \\ 
left wrist & left elbow   & left shoulder   & Left elbow angle \\ 
right wrist & right elbow  & right shoulder  & Right elbow angle \\ 
left knee  & left hip     & left shoulder   & Left hip angle \\ 
right knee & right hip    & right shoulder  & Right hip angle \\ 
left hip   & left knee    & left ankle      & Left knee angle \\ 
right hip  & right knee   & right ankle     & Right knee angle \\ 
\bottomrule
\end{tabular}
\caption{List of angle components used in the study with intuitive explanations.}
\label{tab:angle_pairs_explanation}
\end{table*}

\paragraph{Both} The combination of both coordinates and angles, shown as "both" in the results.

\subsection{Classification Models}
\label{classifier_list}

This work explores three different classification models for GMA, using the extracted time series described in Section~\ref{data_extract}. 5-fold stratified cross-validation (CV) is performed across models to ensure robustness.

\paragraph{1D Convolutional Neural Network (1D-CNN)} 
1D-CNNs are a variant of the more common 2D convolutional neural network, which apply convolution operations to 1D signals, making them well-suited for processing time-series data. These networks have been shown to perform well when working with smaller labelled datasets and for specific applications \citep{kiranyaz20211d}.

\paragraph{Long Short-Term Memory (LSTM)}
LSTM networks \citep{hochreiter1997long} are a type of recurrent neural network capable of exploiting the temporal aspect of our data by learning relevant context information from previously seen time points. LSTMs are ideal for handling time series and assessing whether longer temporal memory helps in recognizing and classifying infant's general movements.

\paragraph{Random Forest (RF)}
RFs \citep{random_forests} are powerful discriminative classifiers that perform well on a variety of datasets with minimal tuning \citep{biau2016random}. It is an ensemble method that builds multiple decision trees trained on a random subset of data and combines their outputs to improve prediction accuracy and reduce overfitting. Although they do not explicitly account for the temporal structure of the data, their efficiency and accuracy make them a strong candidate for our classification tasks.

\section{Experiments, Results and Discussion}
\subsection{Implementation details}
We performed a grid search over the hyperparameter space for each of the three classifier types in Section~\ref{classifier_list}, choosing the values that performed best in the evaluation. The final models were trained using these hyperparameters with 5-fold stratified CV, and the results from the held-out test set, with 20\% of the data, were collected. On the \textit{Label \& Track} approach, the classification models used included a 1D-CNN with binary cross entropy (BCE) loss, learning rate of 0.00001, batch size 6, a fully-connected bottleneck with 150 features, and trained for 150 epochs. The LSTM used the BCE loss, learning rate 0.001, batch size 6, 3 layers, a hidden size of 64 and trained for 200 epochs. Finally, the random forest had 170 estimators. Moreover, using the \textit{Aggpose} approach, the 1D-CNN had a learning rate of 0.0001, batch size of 4 and 150 features in the fully connected bottleck trained for 50 epochs. The LSTM also used BCE loss learning rate 0.0001, batch size 4, 2 layers, a hidden size of 64 and trained for 50 epochs. Lastly, the random forest in this setup had 45 estimators. All experiments were run using 30 independent seeds on each individual set. Our models were implemented using pytorch v2.3.1 and scikit-learn v1.3.0. The final classification was evaluated using accuracy, Area Under the Receiver Operating Characteristic Curve (AUROC), and Area Under the Precision–Recall curve (AUPRC).

\subsection{Results and Discussion}

\begin{table*}
\centering
\footnotesize
\caption{Performance of various classifiers for GM classification across the different keypoint retrieval methods. The mean performance across 30 seeds and 3 keypoint labellers is reported for the \textit{Label \& Track} method, and the mean and standard deviation over 30 seeds for \textit{AggPose} is reported. The best result per model and method is bolded.}
\label{results_table}
\renewcommand{\arraystretch}{1.2}
\resizebox*{!}{\dimexpr\textheight-8\baselineskip\relax}{
\begin{tabular}{c c c c c c }
\toprule
\multirow{2}{*}{\textbf{Metric}} & 
\multirow{2}{*}{\shortstack{\textbf{Keypoint} \textbf{Retrieval}}} & 
\multirow{2}{*}{\shortstack{\textbf{Input} \textbf{Features}}} & 
\multirow{2}{*}{\textbf{RF}} & \multirow{2}{*}{\textbf{LSTM}} & \multirow{2}{*}{\textbf{CNN}} \\
\\
\bottomrule
\bottomrule
\multicolumn{6}{c}{\textbf{Early GM Group}} \\
\toprule

\multirow{6}{*}{\rotatebox[origin=c]{90}{\textbf{AUROC}}} & \multirow{3}{*}{\textit{Label \& Track}} & Coordinates & $0.58$ & $\textbf{0.65}$ & $0.72$ \\
& & Angles & $\textbf{0.77}$ & $0.62$ & $0.75$ \\
& & Both & $0.70$ & $0.65$ & $\textbf{0.75}$ \\
& \multirow{3}{*}{\textit{AggPose}} & Coordinates & $0.48 \pm 0.19$ & $0.57 \pm 0.19$ & $0.46 \pm 0.18$ \\
& & Angles & $\textbf{0.51}\pm 0.19$ & $\textbf{0.61}\pm 0.21$ & $0.49 \pm 0.15$  \\
& & Both & $0.48\pm 0.17$ & $0.53\pm 0.21$ & $\textbf{0.51} \pm 0.19$ \\
\midrule

\multirow{6}{*}{\rotatebox[origin=c]{90}{\textbf{AUPRC}}} & \multirow{3}{*}\textit{{Label \& Track}} & Coordinates & $0.72$ & $\textbf{0.74}$ & $\textbf{0.84}$ \\
& & Angles & $\textbf{0.84}$ & $0.70$ & $0.80$ \\
& & Both & $0.79$ & $0.73$ & $0.84$ \\
& \multirow{3}{*}{\textit{AggPose}} & Coordinates & $0.63 \pm 0.16$ & $0.70\pm 0.15$ & $0.65 \pm 0.15$ \\
& & Angles & $\textbf{0.68}\pm 0.15$ & $\textbf{0.73}\pm 0.16$ & $0.65 \pm 0.13$ \\
& & Both & $0.63\pm 0.16$ & $0.65\pm 0.16$ & $\textbf{0.68} \pm 0.15$ \\
\midrule

\multirow{6}{*}{\rotatebox[origin=c]{90}{\textbf{Accuracy}}} & \multirow{3}{*}{\textit{Label \& Track}} & Coordinates & $0.54$ & $0.60$ & $0.66$ \\
& & Angles & $\textbf{0.69}$ & $0.57$ & $0.67$ \\
& & Both & $0.60$ & $\textbf{0.63}$ & $\textbf{0.67}$ \\
& \multirow{3}{*}{\textit{AggPose}} & Coordinates & $0.47 \pm 0.13$ & $0.54\pm 0.16$ & $0.47 \pm 0.15$ \\
& & Angles & $\textbf{0.50}\pm 0.17$ & $\textbf{0.59}\pm 0.14$ & $0.47 \pm 0.11$ \\
& & Both & $0.47\pm 0.15$ & $0.53\pm 0.20$ & $\textbf{0.49} \pm 0.14$ \\

\bottomrule
\bottomrule
\multicolumn{6}{c}{\textbf{Late GM Group}} \\
\toprule

\multirow{6}{*}{\rotatebox[origin=c]{90}{\textbf{AUROC}}} & \multirow{3}{*}{\textit{Label \& Track}} & Coordinates & $0.45$ & $0.45$ & $\textbf{0.57}$ \\
& & Angles & $\textbf{0.59}$ & $\textbf{0.54}$ & $0.55$ \\
& & Both & $0.48$ & $0.47$ & $0.56$ \\
& \multirow{3}{*}{\textit{AggPose}} & Coordinates & $0.40\pm 0.26$ & $0.54 \pm 0.23$ &$0.56 \pm 0.24$ \\
& & Angles & $\textbf{0.44}\pm 0.25$ & $\textbf{0.66} \pm 0.24$ & $\textbf{0.72} \pm 0.24$ \\
& & Both & $0.40\pm 0.24$ & $0.65 \pm 0.26$ & $0.52 \pm 0.29$ \\
\midrule

\multirow{6}{*}{\rotatebox[origin=c]{90}{\textbf{AUPRC}}} & \multirow{3}{*}{\textit{Label \& Track}} & Coordinates & $0.31$ & $0.36$ & $\textbf{0.50}$ \\
& & Angles & $\textbf{0.45}$ & $\textbf{0.42}$ & $0.39$ \\
& & Both & $0.33$ & $0.35$ & $0.43$ \\
& \multirow{3}{*}{\textit{AggPose}} & Coordinates & $0.36\pm 0.21$ & $0.50 \pm 0.24$ & $0.50 \pm 0.25$ \\
& & Angles & $\textbf{0.36}\pm 0.19$ & $0.58 \pm 0.27$ & $\textbf{0.65} \pm 0.28$ \\
& & Both & $0.33\pm 0.18$ & $\textbf{0.59} \pm 0.28$ & $0.45 \pm 0.28$ \\
\midrule

\multirow{6}{*}{\rotatebox[origin=c]{90}{\textbf{Accuracy}}} & \multirow{3}{*}{\textit{Label \& Track}} & Coordinates & $0.63$ & $0.59$ & $0.60$ \\
& & Angles & $\textbf{0.67}$ & $\textbf{0.64}$ & $0.62$ \\
& & Both & $0.64$ & $0.62$ & $\textbf{0.62}$ \\
& \multirow{3}{*}{\textit{AggPose}} & Coordinates & $0.54\pm 0.17$ & $0.60 \pm 0.17$ & $0.62 \pm 0.18$ \\
& & Angles & $\textbf{0.62}\pm 0.11$ & $\textbf{0.71} \pm 0.15$ & $\textbf{0.70} \pm 0.18$ \\
& & Both & $0.59\pm 0.14$ & $0.61 \pm 0.18$ & $0.61 \pm 0.19$ \\
\bottomrule

\end{tabular}
}
\end{table*}

\autoref{results_table} shows the results of the different classification models, across the different keypoint retrieval methods as measured in AUROC, AUPRC and Accuracy. See Appendix \ref{app:metrics} and \ref{app:label_and_track_vs_aggpose} for a comprehensive overview of the results, which are also broken down by individual keypoint labeller performance. 
They demonstrate a variety of outcomes across age groups and feature sets, with no clear pattern indicating that one method is superior. However, the use of angle features, either isolated or combined, proves useful, as the angles only or both sets of input features generally outperform the coordinates only set, sometimes quite significantly. Additionally, the early GM group shows significantly better overall results, this indicates a large difference in the difficulty of the task between the age groups, given that we have roughly the same number of videos for each age group.  Overall, the \textit{Label \& Track} approach performs comparably to the automatically labeled \textit{AggPose} method. However, in a significant number of data groups, it demonstrates improved results, highlighting the positive impact of keypoint tracking. By incorporating temporal information from the time series to determine the position of keypoints at each time frame, this approach enhances classification performance, whereas \textit{AggPose} labels each keypoint in isolated frames, missing potential contextual cues. Furthermore, the point tracking algorithm used by \textit{Label \& Track} is explicitly trained to handle keypoint occlusions, which occur often in these videos. Finally, the \textit{Label \& Track} approach benefits from the two rounds of manual keypoint correction for detected outliers, as described in Section ~\ref{data_preprocessing}. See Appendix \ref{app:label_and_track_vs_aggpose} for more information regarding the relative performance between \textit{Label \& Track} and \textit{AggPose}, including a breakdown of which experiment settings show statistically significant differences. One of the key takeaways from this study is that despite the relatively small and heterogeneous dataset, our approach has shown success in automatically classifying GM in newborns, with average AUROCs of up to 0.8283 in certain scenarios, opening promising lines of research for GMA ``in the wild''.





\section{Conclusion, Limitations and Future Outlook}
This study presents a preliminary but promising step toward the development of an automated tool for General Movements Assessment (GMA). We demonstrate that infant motor quality can be predicted from video recordings with an AUROC of up to 0.8283, with overall better performance in the early GM group. This highlights the potential of using automated methods to assess infant motor behaviour. Despite the challenges posed by diverse data sources and recording conditions in the data, our results show that automated GMA is feasible in real-world clinical settings. This is a significant step forward in the early detection of neurodevelopmental disorders. This study lays the groundwork for further advancements, emphasizing the need for larger datasets and more refined models and feature extraction approaches to capture the data complexities. Ultimately, the goal is to enhance early prediction of neurodevelopmental diseases, offering a valuable tool for clinicians, and ultimately improving patient care.

\paragraph{Limitations}
One of the key limitations of this study is the small dataset size, which affects both the robustness and generalizability of the results, as reflected in the relatively large standard deviations across cross-validation folds. This also restricts our ability to perform stratified error analyses, such as evaluating performance differences across age groups or recording conditions, both of which are clinically relevant due to the evolving nature of infant movements.

The demographic homogeneity of the dataset is another constraint, particularly in assessing how well the model performs across subpopulations. For example, evaluating the effect of skin tone on keypoint tracking is not feasible with the current cohort.
While our \textit{Label \& Track} approach relies on a tracking algorithm that uses general visual features and is not explicitly tied to skin appearance, this assumed robustness must still be empirically validated. Importantly, the modular nature of our pipeline allows for the integration of improved tracking algorithms as the field advances.

The heterogeneity of the video data (including differences in camera quality, angles, environments, and frame rates) adds complexity to the task but also increases clinical relevance by reflecting real-world variability. However, the limited dataset size again restricts our ability to systematically analyse how these factors influence model performance.

A further limitation is the coarse, video-level annotation of movement quality. In practice, not all segments of a video are equally informative. Since we segment longer recordings into fixed-length clips and treat them independently, the use of a single video-level label introduces noise, especially when only a portion of the video reflects abnormal movements. Finer-grained, segment-level annotations would reduce this label noise and likely improve model accuracy. While such annotations would introduce new challenges (e.g., class imbalance), they would also enable more advanced approaches like Multiple Instance Learning (MIL), which could increase interpretability by focusing on which segments lead to a particular classification. Additionally, since GMA is assessed on a continuous scale, integrating that granularity into modelling could provide richer predictions provided a larger dataset.

Finally, the small dataset also limits our ability to explore more powerful models, such as Vision Transformers (such as TimeSformer), which may better capture the complex spatiotemporal structure of infant movements.


\paragraph{Future Outlook} Future work on this dataset could focus on engineering more sophisticated features from the extracted keypoints particularly those that capture temporal dynamics relevant to infant motor behavior. Exploring more powerful model classes--such as Transformers--may also better leverage the sptial and temporal patterns present in the videos. Addionally, our current outlier detection correction pipeline, used for manually labeled keypoints, could be used to refine the keypoints extracted by the \textit{AggPose} automatic keypoint retrieval method. Replacing the current tracking module in the \textit{Label \& Track} approach with more recent point tracking algorithms could also further improve robustness. 
As the dataset grows, more comprehensive analyses will become possible, including stratified error analysis across different age groups, recording conditions, and demographic subgroups. Interpretability remains a critical goal in medical applications; identifying movement patterns that drive classification decisions would be highly valuable. With more fine-grained or continuous labels, approaches such as Multiple Instance Learning could offer more transparent and clinically relevant predictions. While GMA is a standardized assessment, inter-institutional studies of annotator agreement would further validate model generalizability. 
Ultimately, if predictive performance can be maintained or improved, the development of a decision support tool to assist clinicians in screening more infants using GMA would be a highly impactful application of this work.

\acks{DC is funded from the Strategic Focal Area “Personalized Health and Related Technologies
(PHRT)” grant \#2021-911 of the ETH Domain (Swiss Federal Institutes of Technology), and SL
and KC from the Swiss State Secretariat for Education, Research and Innovation (SERI), contract
\#MB22.00047. 
The authors would like to thank Ri\u cards Marcinkevi\u cs and Heike Leutheuser for their insights and discussions on the development of this manuscript and all the reviewers for their helpful feedback and suggestions.}

\newpage


\bibliography{sample}

\newpage
\appendix
\section{AggPose for Automatic Infant Pose Estimation}
\label{app:aggpose}
To generate automatic keypoint annotations across our video dataset, we employed the \textit{AggPose} framework \citep{cao2022aggpose}, a state-of-the-art vision transformer specifically designed for infant pose estimation. This appendix provides a more detailed overview of the model and our rationale for its use as an automated labelling tool in our pipeline.

\paragraph{Overview and Motivation}
AggPose (Aggregation Vision Transformer) addresses several core challenges in infant pose estimation, including small body sizes, high variability in posture, and the limited availability of large, annotated datasets. Unlike traditional convolutional or hybrid CNN-transformer models, AggPose uses a fully transformer-based architecture that integrates multi-scale spatial information through a deep-layer aggregation mechanism and cross-resolution fusion via multi-layer perceptrons (MLPs). This design enables the model to learn global and local spatial dependencies without relying on convolutional backbones, making it better suited for fine-grained pose analysis in infants.

The model is initially pre-trained on the COCO dataset and subsequently fine-tuned on a dedicated infant dataset introduced by the authors, which includes over 20,000 manually labelled images and millions of unlabeled frames. This domain-specific adaptation allows AggPose to outperform general-purpose pose estimators in capturing the subtleties of infant movement.

\paragraph{Model Architecture}
AggPose is structured around a multi-stage transformer pipeline, where each stage processes feature maps at different spatial resolutions using Mix Transformer (MiT) blocks. These consist of overlapped patch embeddings, multi-head self-attention layers, and feedforward modules (Mix-FFNs) that incorporate depth-wise convolution to improve local detail capture.

To facilitate feature integration across resolutions, the model employs an MLP-based fusion mechanism instead of conventional skip connections. This cross-layer aggregation allows for efficient information sharing across spatial scales, improving convergence speed and robustness. Two model variants are available—AggPose-S and AggPose-L—differing in backbone size and depth. 
According to the original benchmarks, this model achieved 76.4 AP on COCO and 95.0 AP on the infant pose dataset.

\paragraph{Use in Our Study}
We applied AggPose-L to automatically annotate each video frame in our dataset with 21 infant-specific keypoints—extending the COCO format to include additional clinically relevant joints such as fingers, toes, and the navel. These annotations form the basis for downstream analysis of pose and movement over time. The choice of AggPose was motivated by its strong benchmark performance relative to alternatives like OpenPose, HRNet, and TokenPose, as well as its ability to generalize well to infant data.

Beyond raw accuracy, AggPose offers several practical advantages: reliable keypoint detection across varied poses and lighting conditions; consistent labelling across long sequences; and a clinically meaningful keypoint format. Its transformer-based architecture, paired with efficient aggregation, makes it ideal for large-scale, automated annotation tasks requiring both precision and scalability. An example of the extracted keypoints is shown in \autoref{fig:tools_figure}c.

\section{Further Results}
\label{app:metrics}
\autoref{app:auroc_results_table}, \autoref{app:auprc_results_table} and \autoref{app:acc_results_table} show the results of the presented classification models, for each labeller individually as well as averaged, grouped by age and feature extraction method used, measured in AUROC, AUPRC and accuracy, respectively. The \textit{AggPose} results are also included. The best performing method per task is marked in bold.

\begin{table*}
\centering
\footnotesize
\caption{\textbf{AUROC} of various classifiers for GM classification across different keypoint labellers. The mean and standard deviation over 30 seeds is reported. Best result per model and labeller is bolded.}
\label{app:auroc_results_table}
\renewcommand{\arraystretch}{1.2}
\resizebox*{!}{\dimexpr\textheight-8\baselineskip\relax}{
\begin{tabular}{c c c c c }
\toprule
\textbf{Keypoint Retrieval} & \textbf{Input Features} & \textbf{RF} & \textbf{LSTM} & \textbf{CNN} \\
\bottomrule
\bottomrule
\multicolumn{5}{c}{\textbf{Early GM Group}} \\
\toprule

& Coordinates & $0.66 \pm 0.19$ & $0.64 \pm 0.16$ & $0.76 \pm 0.19$ \\
\shortstack{\textit{Label \& Track}\\ Labeller \#1} & Angles & $\textbf{0.83} \pm \textbf{0.13}$ & $0.66 \pm 0.16$ & $\textbf{0.82} \pm \textbf{0.14}$ \\
& Both & $0.79 \pm 0.14$ & $\textbf{0.67} \pm \textbf{0.16}$ & $0.79 \pm 0.17$ \\
\midrule

& Coordinates & $0.49 \pm 0.19$ & $\textbf{0.68} \pm \textbf{0.17}$ & $0.65 \pm 0.19$ \\
\shortstack{\textit{Label \& Track}\\ Labeller \#2} & Angles & $\textbf{0.68} \pm \textbf{0.15}$ & $0.63 \pm 0.17$ & $0.61 \pm 0.19$ \\
& Both & $0.57 \pm 0.20$ & $0.61 \pm 0.16$ & $\textbf{0.67} \pm \textbf{0.20}$ \\
\midrule

& Coordinates & $0.59 \pm 0.21$ & $0.62 \pm 0.18$ & $0.75 \pm 0.20$ \\
\shortstack{\textit{Label \& Track}\\ Labeller \#3} & Angles & $\textbf{0.81} \pm \textbf{0.12}$ & $0.56 \pm 0.16$ & $\textbf{0.81} \pm \textbf{0.14}$ \\
& Both & $0.73 \pm 0.18$ & $\textbf{0.66} \pm \textbf{0.21}$ & $0.78 \pm 0.17$ \\
\midrule

& Coordinates & $0.58$ & $\textbf{0.65}$ & $0.72$ \\
\shortstack{\textit{Label \& Track}\\Mean} & Angles & $\textbf{0.77}$ & $0.62$ & $0.75$ \\
& Both & $0.70$ & $0.65$ & $\textbf{0.75}$ \\
\midrule

& Coordinates & $0.48 \pm 0.19$ & $0.57 \pm 0.19$ & $0.46 \pm 0.18$ \\
\textit{AggPose} & Angles & $\textbf{0.51}\pm 0.19$ & $\textbf{0.61}\pm 0.21$ & $0.49 \pm 0.15$ \\
& Both & $0.48\pm 0.17$ & $0.53\pm 0.21$ & $\textbf{0.51} \pm 0.19$ \\

\bottomrule
\bottomrule
\multicolumn{5}{c}{\textbf{Late GM Group}} \\
\toprule

& Coordinates & $0.37 \pm 0.25$ & $0.44 \pm 0.23$ & $\textbf{0.58} \pm \textbf{0.23}$ \\
\shortstack{\textit{Label \& Track}\\ Labeller \#1} & Angles & $\textbf{0.51} \pm \textbf{0.25}$ & $\textbf{0.61} \pm \textbf{0.21}$ & $0.38 \pm 0.22$ \\
& Both & $0.38 \pm 0.24$ & $0.40 \pm 0.19$ & $0.54 \pm 0.22$ \\
\midrule

& Coordinates & $0.45 \pm 0.24$ & $0.49 \pm 0.24$ & $0.51 \pm 0.20$ \\
\shortstack{\textit{Label \& Track}\\ Labeller \#2} & Angles & $\textbf{0.63} \pm \textbf{0.20}$ & $0.49 \pm 0.22$ & $\textbf{0.69} \pm \textbf{0.22}$ \\
& Both & $0.49 \pm 0.23$ & $\textbf{0.55} \pm \textbf{0.18}$ & $0.59 \pm 0.20$ \\
\midrule

& Coordinates & $0.52 \pm 0.27$ & $0.43 \pm 0.27$ & $0.62 \pm 0.19$ \\
\shortstack{\textit{Label \& Track}\\ Labeller \#3} & Angles & $\textbf{0.63} \pm \textbf{0.22}$ & $\textbf{0.54} \pm \textbf{0.23}$ & $\textbf{0.59} \pm \textbf{0.24}$ \\
& Both & $0.57 \pm 0.18$ & $0.48 \pm 0.25$ & $0.54 \pm 0.29$ \\
\midrule

& Coordinates & $0.45$ & $0.45$ & $\textbf{0.57}$ \\
\shortstack{\textit{Label \& Track}\\Mean} & Angles & $\textbf{0.59}$ & $\textbf{0.54}$ & $0.55$ \\
& Both & $0.48$ & $0.47$ & $0.56$ \\
\midrule

& Coordinates & $0.40\pm 0.26$ & $0.54 \pm 0.23$ &$0.56 \pm 0.24$ \\
\textit{AggPose} & Angles & $\textbf{0.44}\pm 0.25$ & $\textbf{0.66} \pm 0.24$ & $\textbf{0.72} \pm 0.24$ \\
& Both & $0.40\pm 0.24$ & $0.65 \pm 0.26$ & $0.52 \pm 0.29$ \\
\bottomrule

\end{tabular}
}
\end{table*}

\begin{table*}
\centering
\footnotesize
\caption{Results. \textbf{AUPRC} over 30 seeds is reported. The best result for each model and labeller is bolded.}
\label{app:auprc_results_table}
\renewcommand{\arraystretch}{1.2}
\resizebox*{!}{\dimexpr\textheight-8\baselineskip\relax}{
\begin{tabular}{c c c c c }
\toprule
\textbf{Keypoint Retrieval} & \textbf{Input Features} & \textbf{RF} & \textbf{LSTM} & \textbf{CNN} \\
\bottomrule
\bottomrule
\multicolumn{5}{c}{\textbf{Early GM Group}} \\
\toprule

& Coordinates & $0.77 \pm 0.14$ & $\textbf{0.75} \pm \textbf{0.13}$ & $0.86 \pm 0.11$ \\
\shortstack{\textit{Label \& Track}\\ Labeller \#1} & Angles & $\textbf{0.89} \pm \textbf{0.09}$ & $0.73 \pm 0.16$ & $\textbf{0.88} \pm \textbf{0.10}$ \\
& Both & $0.86 \pm 0.09$ & $0.74 \pm 0.15$ & $0.85 \pm 0.13$ \\
\midrule
& Coordinates & $0.67 \pm 0.14$ & $\textbf{0.78} \pm \textbf{0.14}$ & $\textbf{0.80} \pm \textbf{0.12}$ \\
\shortstack{\textit{Label \& Track}\\ Labeller \#2} & Angles & $\textbf{0.77} \pm \textbf{0.13}$ & $0.71 \pm 0.17$ & $0.69 \pm 0.16$ \\
& Both & $0.71 \pm 0.15$ & $0.71 \pm 0.15$ & $0.80 \pm 0.14$ \\
\midrule
& Coordinates & $0.73 \pm 0.15$ & $0.70 \pm 0.17$ & $0.85 \pm 0.13$ \\
\shortstack{\textit{Label \& Track}\\ Labeller \#3} & Angles & $\textbf{0.87} \pm \textbf{0.09}$ & $0.67 \pm 0.16$ & $0.85 \pm 0.14$ \\
& Both & $0.81 \pm 0.13$ & $\textbf{0.75} \pm \textbf{0.17}$ & $\textbf{0.86} \pm \textbf{0.12}$ \\
\midrule 
& Coordinates & $0.72$ & $\textbf{0.74}$ & $\textbf{0.84}$ \\
\shortstack{\textit{Label \& Track}\\Mean} & Angles & $\textbf{0.84}$ & $0.70$ & $0.80$ \\
& Both & $0.79$ & $0.73$ & $0.84$ \\
\midrule

& Coordinates & $0.63 \pm 0.16$ & $0.70\pm 0.15$ & $0.65 \pm 0.15$ \\
\textit{AggPose} & Angles & $\textbf{0.68}\pm 0.15$ & $\textbf{0.73}\pm 0.16$ & $0.65 \pm 0.13$ \\
& Both & $0.63\pm 0.16$ & $0.65\pm 0.16$ & $\textbf{0.68} \pm 0.15$ \\

\bottomrule
\bottomrule
\multicolumn{5}{c}{\textbf{Late GM Group}} \\
\toprule

& Coordinates & $0.26 \pm 0.14$ & $0.36 \pm 0.22$ & $\textbf{0.46} \pm \textbf{0.23}$ \\
\shortstack{\textit{Label \& Track}\\ Labeller \#1} & Angles & $\textbf{0.39} \pm \textbf{0.25}$ & $\textbf{0.47} \pm \textbf{0.24}$ & $0.27 \pm 0.17$ \\
& Both & $0.29 \pm 0.19$ & $0.29 \pm 0.15$ & $0.40 \pm 0.22$ \\
\midrule
& Coordinates & $0.31 \pm 0.18$ & $0.38 \pm 0.27$ & $0.48 \pm 0.24$ \\
\shortstack{\textit{Label \& Track}\\ Labeller \#2} & Angles & $\textbf{0.47} \pm \textbf{0.25}$ & $\textbf{0.38} \pm \textbf{0.22}$ & $\textbf{0.50} \pm \textbf{0.27}$ \\
& Both & $0.35 \pm 0.18$ & $0.38 \pm 0.22$ & $0.46 \pm 0.25$ \\
\midrule
& Coordinates & $0.36 \pm 0.19$ & $0.34 \pm 0.23$ & $\textbf{0.55} \pm \textbf{0.21}$ \\
\shortstack{\textit{Label \& Track}\\ Labeller \#3} & Angles & $\textbf{0.49} \pm \textbf{0.28}$ & $\textbf{0.42} \pm \textbf{0.24}$ & $0.41 \pm 0.25$ \\
& Both & $0.36 \pm 0.18$ & $0.38 \pm 0.22$ & $0.43 \pm 0.27$ \\
\midrule 
& Coordinates & $0.31$ & $0.36$ & $\textbf{0.50}$ \\
\shortstack{\textit{Label \& Track}\\Mean} & Angles & $\textbf{0.45}$ & $\textbf{0.42}$ & $0.39$ \\
& Both & $0.33$ & $0.35$ & $0.43$ \\
\midrule

& Coordinates & $0.36\pm 0.21$ & $0.50 \pm 0.24$ & $0.50 \pm 0.25$ \\
\textit{AggPose} & Angles & $\textbf{0.36}\pm 0.19$ & $0.58 \pm 0.27$ & $\textbf{0.65} \pm 0.28$ \\
& Both & $0.33\pm 0.18$ & $\textbf{0.59} \pm 0.28$ & $0.45 \pm 0.28$ \\

\bottomrule
\end{tabular}}
\end{table*}

\begin{table*}
\centering
\footnotesize
\caption{Results. \textbf{Accuracy} over 30 seeds is reported. The best result for each model and labeller is bolded.}
\label{app:acc_results_table}
\renewcommand{\arraystretch}{1.2}
\resizebox*{!}{\dimexpr\textheight-8\baselineskip\relax}{
\begin{tabular}{c c c c c }
\toprule
\textbf{Keypoint Retrieval} & \textbf{Input Features} & \textbf{RF} & \textbf{LSTM} & \textbf{CNN} \\
\bottomrule
\bottomrule
\multicolumn{5}{c}{\textbf{Early GM Group}} \\
\toprule
& Coordinates & $0.60 \pm 0.16$ & $0.59 \pm 0.15$ & $0.69 \pm 0.15$ \\
\shortstack{\textit{Label \& Track}\\ Labeller \#1}  & Angles & $\textbf{0.73} \pm \textbf{0.13}$ & $0.60 \pm 0.14$ & $\textbf{0.70} \pm \textbf{0.16}$ \\
& Both & $0.66 \pm 0.15$ & $\textbf{0.64} \pm \textbf{0.16}$ & $0.68 \pm 0.12$ \\
\midrule
& Coordinates & $0.45 \pm 0.17$ & $\textbf{0.60} \pm \textbf{0.15}$ & $0.62 \pm 0.13$ \\
\shortstack{\textit{Label \& Track}\\ Labeller \#2} & Angles & $\textbf{0.65} \pm \textbf{0.14}$ & $0.56 \pm 0.14$ & $0.56 \pm 0.17$ \\
& Both & $0.53 \pm 0.18$ & $0.59 \pm 0.16$ & $\textbf{0.63} \pm \textbf{0.15}$ \\
\midrule
& Coordinates & $0.55 \pm 0.18$ & $0.61 \pm 0.14$ & $0.68 \pm 0.14$ \\
\shortstack{\textit{Label \& Track}\\ Labeller \#3} & Angles & $\textbf{0.70} \pm \textbf{0.11}$ & $0.54 \pm 0.15$ & $\textbf{0.75} \pm \textbf{0.12}$ \\
& Both & $0.62 \pm 0.17$ & $\textbf{0.65} \pm \textbf{0.17}$ & $0.71 \pm 0.14$ \\
\midrule 
& Coordinates & $0.54$ & $0.60$ & $0.66$ \\
\shortstack{\textit{Label \& Track}\\Mean} & Angles & $\textbf{0.69}$ & $0.57$ & $0.67$ \\
& Both & $0.60$ & $\textbf{0.63}$ & $\textbf{0.67}$ \\
\midrule

& Coordinates & $0.47 \pm 0.13$ & $0.54 \pm 0.16$ & $0.47 \pm 0.15$ \\
\textit{AggPose} & Angles & $\textbf{0.50} \pm 0.17$ & $\textbf{0.59} \pm 0.14$ & $0.47 \pm 0.11$ \\
& Both & $0.47 \pm 0.15$ & $0.53 \pm 0.20$ & $\textbf{0.49} \pm 0.14$ \\

\bottomrule
\bottomrule
\multicolumn{5}{c}{\textbf{Late GM Group}} \\
\toprule

& Coordinates & $0.58 \pm 0.14$ & $0.58 \pm 0.11$ & $0.61 \pm 0.16$ \\
\shortstack{\textit{Label \& Track}\\ Labeller \#1}  & Angles & $\textbf{0.63} \pm \textbf{0.17}$ & $\textbf{0.68} \pm \textbf{0.13}$ & $0.57 \pm 0.13$ \\
& Both & $0.61 \pm 0.16$ & $0.58 \pm 0.16$ & $\textbf{0.62} \pm \textbf{0.15}$ \\
\midrule
& Coordinates & $0.64 \pm 0.13$ & $0.59 \pm 0.15$ & $0.56 \pm 0.21$ \\
\shortstack{\textit{Label \& Track}\\ Labeller \#2} & Angles & $\textbf{0.68} \pm \textbf{0.08}$ & $0.63 \pm 0.14$ & $\textbf{0.64} \pm \textbf{0.13}$ \\
& Both & $0.64 \pm 0.13$ & $\textbf{0.64} \pm \textbf{0.14}$ & $0.61 \pm 0.16$ \\
\midrule
& Coordinates & $0.67 \pm 0.14$ & $0.60 \pm 0.15$ & $0.61 \pm 0.19$ \\
\shortstack{\textit{Label \& Track}\\ Labeller \#3} & Angles & $\textbf{0.70} \pm \textbf{0.08}$ & $0.62 \pm 0.16$ & $0.64 \pm 0.12$ \\
& Both & $0.66 \pm 0.09$ & $\textbf{0.65} \pm \textbf{0.17}$ & $\textbf{0.64} \pm \textbf{0.13}$ \\
\midrule 
& Coordinates & $0.63$ & $0.59$ & $0.60$ \\
\shortstack{\textit{Label \& Track}\\Mean} & Angles & $\textbf{0.67}$ & $\textbf{0.64}$ & $0.62$ \\
& Both & $0.64$ & $0.62$ & $\textbf{0.62}$ \\
\midrule

& Coordinates & $0.54 \pm 0.17$ & $0.60 \pm 0.17$ & $0.62 \pm 0.18$ \\
\textit{AggPose} & Angles & $\textbf{0.62} \pm 0.11$ & $\textbf{0.71} \pm 0.15$ & $\textbf{0.70} \pm 0.18$ \\
& Both & $0.59 \pm 0.14$ & $0.61 \pm 0.18$ & $0.61 \pm 0.19$ \\

 \bottomrule
\end{tabular}}
\end{table*}

\newpage

\section{Label \& Track vs. AggPose}
\label{app:label_and_track_vs_aggpose}

\autoref{relative_performance_table} shows the difference in performance between the \textit{Label \& Track} and \textit{AggPose} methods for keypoint tracking, statistically significant (p $<$ 0.05) differences are bolded. For comparisons where one of the two methods is not normally distributed (according to the Shapiro-Wilk test), the Mann-Whitney U Test was used, otherwise a t-test was used. \autoref{relative_performance_table_pvalues} shows the calculated p values.
\begin{table*}[h!]
\centering
\footnotesize
\caption{Difference in performance of the Label \& Track method (mean over all labellers) vs. AggPose, bolded differences are statistically significant (p $<$ 0.05), p-values are given in the following table. Values marked with an asterisk were calculated using a t-test, all other values used the Mann-Whitney U test.}
\label{relative_performance_table}
\renewcommand{\arraystretch}{1.2}
\resizebox{\textwidth}{!}{
\begin{tabular}{c l c c c c c c}
\toprule
\multirow{2}{*}{\textbf{Metric}} & 
\multirow{2}{*}{\shortstack{\textbf{Input} \\ \textbf{Features}}} & 
\multicolumn{3}{c}{\textbf{Early GM Group}} & 
\multicolumn{3}{c}{\textbf{Late GM Group}} \\
\cmidrule(lr){3-5} \cmidrule(lr){6-8}
& & \textbf{RF} & \textbf{LSTM} & \textbf{CNN} & \textbf{RF} & \textbf{LSTM} & \textbf{CNN} \\
\midrule

\multirow{5}{*}{\rotatebox[origin=c]{90}{\textbf{AUROC}}} 
\\
& Coord. & $\textbf{0.1005*}$ & $\textbf{0.1226*}$ & $\textbf{0.2578}$ & $0.0463$ & $-0.0889*$ & $0.0071$ \\
& Angles & $\textbf{0.2641}$ & $0.0075*$ & $\textbf{0.2546}$ & $\textbf{0.1511}$ & $\textbf{-0.1146*}$ & $\textbf{-0.1625}$ \\
& Both & $\textbf{0.2122}$ & $\textbf{0.1091*}$ & $\textbf{0.2383}$ & $0.0735*$ & $\textbf{-0.1737*}$ & $0.0342*$ \\
\\
\midrule

\multirow{5}{*}{\rotatebox[origin=c]{90}{\textbf{AUPRC}}} 
\\
& Coord. & $\textbf{0.0890*}$ & $\textbf{0.0954}$ & $\textbf{0.1865}$ & $-0.0452$ & $\textbf{-0.1412}$ & $0.0008$ \\
& Angles & $\textbf{0.1623}$ & $-0.0290$ & $\textbf{0.1500}$ & $0.0878$ & $\textbf{-0.1529}$ & $\textbf{-0.2629}$ \\
& Both & $\textbf{0.1587}$ & $\textbf{0.0857}$ & $\textbf{0.1591}$ & $-0.0005$ & $\textbf{-0.2397}$ & $-0.0207$ \\
\\
\midrule

\multirow{5}{*}{\rotatebox[origin=c]{90}{\textbf{Accuracy}}} 
\\
& Coord. & $0.0671*$ & $\textbf{0.0540}$ & $\textbf{0.1980}$ & $\textbf{0.0939}$ & $-0.0122*$ & $-0.0277$ \\
& Angles & $\textbf{0.1912*}$ & $-0.0235*$ & $\textbf{0.2000*}$ & $\textbf{0.0504}$ & $\textbf{-0.0693*}$ & $\textbf{-0.0876}$ \\
& Both & $\textbf{0.1271}$ & $\textbf{0.0987*}$ & $\textbf{0.1863}$ & $0.0426$ & $0.0143*$ & $0.0098$ \\
\\

\bottomrule
\end{tabular}
}
\end{table*}

\newpage
\begin{table*}[h!]
\centering
\footnotesize
\caption{P values for relative performance of the Label \& Track method vs. AggPose, bolded differences are statistically significant (p $<$ 0.05). Values marked with an asterisk were calculated using a t-test, all other values used the Mann-Whitney U test.}
\label{relative_performance_table_pvalues}
\renewcommand{\arraystretch}{1.2}
\resizebox{\textwidth}{!}{
\begin{tabular}{c l c c c c c c}
\toprule
\multirow{2}{*}{\textbf{Metric}} & 
\multirow{2}{*}{\shortstack{\textbf{Input} \\ \textbf{Features}}} & 
\multicolumn{3}{c}{\textbf{Early GM Group}} & 
\multicolumn{3}{c}{\textbf{Late GM Group}} \\
\cmidrule(lr){3-5} \cmidrule(lr){6-8}
& & \textbf{RF} & \textbf{LSTM} & \textbf{CNN} & \textbf{RF} & \textbf{LSTM} & \textbf{CNN} \\
\midrule

\multirow{5}{*}{\rotatebox[origin=c]{90}{\textbf{AUROC}}} 
\\
& Coord. & $\textbf{0.0213*}$ & $\textbf{0.0016*}$ & $\textbf{0.0000}$ & $0.3728$ & $0.0908*$ & $0.9536$ \\
& Angles & $\textbf{0.0000}$ & $0.8431*$ & $\textbf{0.0000}$ & $\textbf{0.0075}$ & $\textbf{0.0192*}$ & $\textbf{0.0051}$ \\
& Both & $\textbf{0.0000}$ & $\textbf{0.0069*}$ & $\textbf{0.0000}$ & $0.1381*$ & $\textbf{0.0004*}$ & $0.5183*$ \\
\\
\midrule

\multirow{5}{*}{\rotatebox[origin=c]{90}{\textbf{AUPRC}}} 
\\
& Coord. & $\textbf{0.0070*}$ & $\textbf{0.0040}$ & $\textbf{0.0000}$ & $0.3200$ & $\textbf{0.0030}$ & $0.8771$ \\
& Angles & $\textbf{0.0000}$ & $0.4397$ & $\textbf{0.0000}$ & $0.1813$ & $\textbf{0.0040}$ & $\textbf{0.0000}$ \\
& Both & $\textbf{0.0000}$ & $\textbf{0.0107}$ & $\textbf{0.0000}$ & $1.0000$ & $\textbf{0.0000}$ & $0.7548$ \\
\\
\midrule

\multirow{5}{*}{\rotatebox[origin=c]{90}{\textbf{Accuracy}}} 
\\
& Coord. & $0.0547*$ & $\textbf{0.0405}$ & $\textbf{0.0000}$ & $\textbf{0.0048}$ & $0.6917*$ & $0.8148$ \\
& Angles & $\textbf{0.0000*}$ & $0.4485*$ & $\textbf{0.0000*}$ & $\textbf{0.0325}$ & $\textbf{0.0248*}$ & $\textbf{0.0071}$ \\
& Both & $\textbf{0.0008}$ & $\textbf{0.0072*}$ & $\textbf{0.0000}$ & $0.0742$ & $0.6787*$ & $0.6325$ \\
\\

\bottomrule
\end{tabular}
}
\end{table*}

\end{document}